\documentclass{article}

\usepackage{arxiv}

\usepackage[utf8]{inputenc}
\usepackage[T1]{fontenc}
\usepackage{hyperref}
\usepackage{url}
\usepackage{booktabs}
\usepackage{amsfonts}
\usepackage{nicefrac}
\usepackage{microtype}
\usepackage{lipsum}
\usepackage{graphicx}
\usepackage{natbib}
\usepackage{pifont}
\usepackage{amsmath}
\usepackage{float}
\usepackage{booktabs}  
\usepackage{caption}

\usepackage{tikz}
\usetikzlibrary{positioning,arrows.meta}
\usepackage{adjustbox}
\usepackage{siunitx}

\makeatletter
\renewcommand{\@date}{}
\makeatother

\title{HAEPO: History-Aggregated Exploratory Policy Optimization}

\date{}

\author{
  Gaurish Trivedi\thanks{Equal contribution. All emails use the domain \texttt{@pilani.bits-pilani.ac.in}.} \\
  Birla Institute of Technology and Science, Pilani\\
  Pilani, Rajasthan (333031) \\
  \texttt{f20220728} 
  \And
  Alakh Sharma\footnotemark[1] \\
  Birla Institute of Technology and Science, Pilani\\
  Pilani, Rajasthan (333031) \\
  \texttt{f20240593} 
  \And
  Kartikey Singh Bhandari \\
  Birla Institute of Technology and Science, Pilani\\
  Pilani, Rajasthan (333031) \\
  \texttt{p20241006} 
  \And
  Dhruv Kumar \\
  Birla Institute of Technology and Science, Pilani\\
  Pilani, Rajasthan (333031) \\
  \texttt{dhruv.kumar} 
  \And
  Pratik Narang \\
  Birla Institute of Technology and Science, Pilani\\
  Pilani, Rajasthan (333031) \\
  \texttt{pratik.narang} 
  \And
  Jagat Sesh Challa \\
  Birla Institute of Technology and Science, Pilani\\
  Pilani, Rajasthan (333031) \\
  \texttt{jagatshesh} 
}

\begin{document}
\maketitle
\begin{abstract}
Exploration is essential in modern learning, from reinforcement learning environments with small neural policies to large language models (LLMs). Existing work, such as DPO, leverages full sequence log‑likelihoods to capture an entire trajectory of the model's decisions, while methods like GRPO aggregate per‑token ratios into a trajectory‑level update. However, both often limit exploration on long-horizon tasks. We introduce History-Aggregated Exploratory Policy Optimization (HAEPO), a history‑aware exploratory loss to combat these shortcomings. HAEPO compresses each trajectory into the sum of its logarithmic probabilities (a cumulative logarithmic likelihood), and applies a Plackett-Luce softmax across trajectories to obtain normalized weights proportional to their returns, thus encouraging broader exploration. We add entropy regularization to stabilize the aggressive updates to prevent premature collapse and a soft KL penalty relative to a frozen copy of the previous (reference) policy. Empirically, HAEPO converges fast, explores thoroughly, aligns closely with true rewards, and demonstrates robust learning behavior better or at par with PPO, GRPO, and DPO across diverse tasks. Thus, HAEPO provides a stable and interpretable framework by explicitly leveraging full‑trajectory history while balancing exploration and stability.
\end{abstract}

\section{Introduction}
Reinforcement learning (RL) is a framework in which an agent interacts sequentially with an environment to maximize cumulative scalar rewards over time \citep{sutton2018reinforcement}.  
Balancing exploration (trying new actions) and exploitation (leveraging known high‑reward behaviors) is fundamental for discovering optimal policies \citep{schulman2015high}.  
Policy‑gradient methods optimize a parameterized policy by computing gradients of the expected return concerning policy parameters, updating via the log‑probability of chosen actions \citep{williams1992simple}.  
While powerful in continuous action domains, these per‑step estimators often suffer from high variance and poor credit assignment in long‑horizon tasks \citep{schulman2015high}. 

To mitigate variance, baselines and generalized advantage estimation (GAE) subtract a value estimate from the returns at each timestep \citep{schulman2015high}.  
However, for tasks that span hundreds or thousands of steps, per‑step gradients still struggle to capture delayed reward structures effectively \citep{bellemare2016unifying}.  
In response, trajectory‑level approaches aggregate log‑probabilities across entire episodes into a single optimization objective, naturally capturing long‑term dependencies \citep{williams1992simple}.  
Works such as RLHF employ PPO to fine‑tune LLMs by summing sequence‑level log‑likelihoods weighted by human preference rewards \citep{ouyang2022traininglanguagemodelsfollow}. However, these trajectory-level methods still suffer from unnormalized trajectory weights (high gradient variance), per-step clipping (restricted exploration), and the loss of full list-wise ranking information.

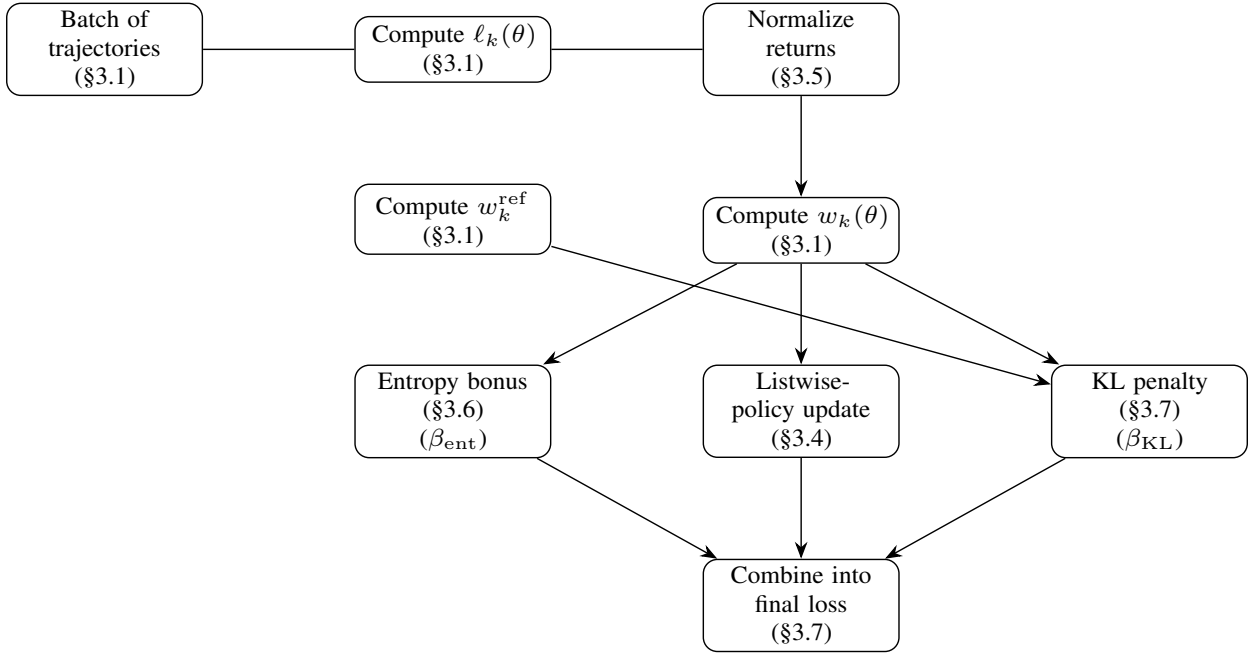
\begin{figure}[H]
  \centering
  \begin{adjustbox}{width=\columnwidth,center}
    \begin{tikzpicture}[
        ->, >=Stealth,
        node distance=1cm and 1.5cm,
        font=\scriptsize,
        block/.style={
          rectangle, draw, rounded corners, align=center,
          text width=1.8cm, inner sep=2pt
        }
      ]
      \node[block]            (A) {Batch of\\trajectories\\(\S \ref{sec:traj-ll})};
      \node[block,right=of A] (B) {Compute $\ell_k(\theta)$\\(\S \ref{sec:traj-ll})};
      \node[block,right=of B] (C) {Normalize returns\\(\S \ref{sec:reward-norm})};

      \node[block,below=of C] (D) {Compute $w_k(\theta)$\\(\S \ref{sec:traj-ll})};
      \node[block,below=of B] (E) {Compute $w_k^{\rm ref}$\\(\S \ref{sec:traj-ll})};

      \node[block,below left=of D]  (F) {Entropy bonus\\(\S \ref{sec:entropy-plpo})\\($\beta_{\rm ent}$)};
      \node[block,below=of D]       (G) {Listwise-policy update\\(\S \ref{sec:orig-loss})};
      \node[block,below right=of D] (H) {KL penalty\\(\S \ref{sec:kl-plpo})\\($\beta_{\rm KL}$)};

      \node[block,below=of G]       (I) {Combine into\\final loss\\(\S \ref{sec:kl-plpo})};

      \draw (A) -- (B) -- (C) -- (D);
      \draw (D) -- (F);
      \draw (D) -- (G);
      \draw (D) -- (H);
      \draw (E) -- (H);
      \draw (F) -- (I);
      \draw (G) -- (I);
      \draw (H) -- (I);
    \end{tikzpicture}
  \end{adjustbox}
  \caption[HAEPO workflow]{Loss-function workflow for History‑Aggregated Exploratory Policy Optimization (HAEPO) (\S \ref{sec:pl-model}).}  
  \label{fig:haepo_workflow}
\end{figure}

In order to overcome the gaps in the existing work, we propose \textit{History‑Aggregated Exploratory Policy Optimization} \textbf{(HAEPO)} in this paper. HAEPO (\S \ref{sec:pl-model} and Fig.~\ref{fig:haepo_workflow}) applies a Plackett-Luce softmax over the sum of log‑probabilities of each trajectory, weighting trajectories proportionally to their normalized returns \citep{plackett1975analysis}.  
We then combine this list-wise weighting with entropy regularization and a soft KL penalty to a frozen reference policy, ensuring robust exploration and stable convergence.  
Conceptually, unlike DPO and GRPO, which focus on per‑token or per‑sequence ratios, HAEPO’s listwise normalization balances diverse high‑return trajectories, improving exploration in sparse and long‑horizon settings \citep{Schulman2017ProximalPO}.  More specifically, HAEPO provides the following advantages: \ding{182} \textbf{maximizing exploration} by amplifying diverse, high‑return trajectories through Plackett-Luce weighting; \ding{183} \textbf{high update efficiency} by keeping each policy update cheap, enabling many iterations per unit time; \ding{184} \textbf{time‑efficient exploration} by converting fast iterations into broader state‑action coverage per wall‑clock minute; \ding{185} \textbf{resource frugality} by minimizing GPU memory and compute demands for scalable long‑horizon training.

Empirical results demonstrate faster convergence and lower variance on bandit and Random Walk benchmarks. On the classic 30‑armed bandit, HAEPO converges faster and with fewer fluctuations than DPO and PPO, owing to its trajectory‑level weighting and stabilization terms.  
In CartPole, HAEPO matches PPO’s final performance while delivering higher update throughput and greater exploration rate per second of training.  
In human‑feedback fine‑tuning, HAEPO performs on par with GRPO while reducing GPU memory usage by 26.4\%, which is comparable to DPO, hence enabling stable, sample‑efficient learning within practical hardware constraints.

\section{Related Work}
\subsection{Exploration Strategies and Trajectory-Level Methods}
Compelling exploration influences performance in RL and LLMs' fine-tuning tasks. Achbany et al. \citep{Achbany2008TuningCE} present an optimal exploration strategy leveraging Boltzmann distributions over Q-values to balance exploration and exploitation. Hao et al. \citep{Yang2021ExplorationID} survey exploration methodologies, categorizing them into uncertainty-oriented and intrinsic motivation-oriented approaches, emphasizing difficulties such as sparse rewards and prolonged horizons. Methods employing full-trajectory likelihoods to guide exploration include variational policy search \citep{Levine2013VariationalPS}, latent trajectory optimization \citep{Luck2019ImprovedET}, and planning methods that maximize expected information gain \citep{Mehta2022ExplorationVP}. However, these approaches lose informative distinctions among sequences \citep{Ke2019LearningDM,Pitis2020MaximumEG}. Token-level ratio aggregation techniques, exemplified by policy search methods in robotics \citep{deisenroth2013survey,Tang2024DeepRL}, aggregate per-token probabilities into trajectory-level updates but often dilute exploration signals for long-horizon tasks. Challenges like credit assignment and variance explosion \citep{Nair2017OvercomingEI}, sparse-reward navigation \citep{Pitis2020MaximumEG}, and unstable multi-agent dynamics \citep{Yang2021ExplorationID} complicate exploration scaling. \textit{Despite advances in full-trajectory exploration, existing methods either collapse subtle sequence distinctions into coarse planning objectives or dilute long-horizon signals through per-token aggregation, limiting credit assignment and stability. We overcome this gap in our proposed method by compressing each rollout into a single cumulative log-likelihood and then applying a Plackett-Luce softmax over this cumulative, so that weights preserve every fine-grained sequence distinction and amplify long-horizon exploration signals, rather than diluting them via per-token aggregation}.

\subsection{Stabilization via Regularization and Trust-Region Constraints}
Entropy-based regularization smoothens the loss landscape to enable stable and robust policy updates \citep{Ahmed2018UnderstandingTI, Brekelmans2022YourPR}. Mirror Descent projects onto a high-entropy simplex to avoid collapse \citep{Neu2017AUV}. Sample-aware entropy regularization has further improved off-policy stability \citep{Han2020DiversityAS}. However, excessive entropy regularization compromises convergence speed by risking overly stochastic policies. Trust-region methods offer an alternative stability mechanism by penalizing policy divergence from a reference policy. Examples such as Constrained Policy Optimization \citep{Achiam2017ConstrainedPO}, Projection-Based Constrained Policy Optimization \citep{Narasimhan2020ProjectionBasedCP}, and Penalized Proximal Policy Optimization \citep{Zhang2022PenalizedPP} adopt this. \textit{Although these methods successfully constrain policy updates, rigid constraints can unduly restrict exploration. We overcome this gap in our proposed method by adding both an entropy bonus for spread-out exploration and a soft KL penalty against a frozen policy, thus effectively enforcing a trust-region-style constraint on the trajectory-centric loss.}

\subsection{Preference-Optimization for LLM Alignment}
Various \textit{pairwise preference-optimization} algorithms align LLMs with human feedback. (1) Direct Preference Optimization (DPO) \citep{rafailov2023direct} bypasses explicit reward modeling by reparameterizing the implicit reward into a simple classification loss over human comparisons, matching or exceeding PPO‐based RLHF in summarization and dialogue tasks while simplifying implementation and training. (2) Kahneman-Tversky Optimization (KTO) by \citep{ethayarajh2024kto} builds upon the cognitive decision theory, which integrates prospect-theoretic utility functions to capture biases like loss aversion, optimizing LLMs using only a binary “good vs. bad” signal across model scales from 1 B to 30 B parameters. (3) Odds Ratio Preference Optimization (ORPO) \citep{hong2024orpo} streamlines alignment by embedding a log‐odds penalty directly into the supervised fine‐tuning loss, eliminating the need for a separate reference model and demonstrating strong performance on AlpacaEval and MT‐Bench. In contrast to \textit{pairwise methods}, \textit{group‐relative} and \textit{listwise approaches} offer richer optimization paradigms. (4) Group Relative Policy Optimization (GRPO) \citep{shao2024deepseekmath} adapts the PPO surrogate by replacing the learned critic with a baseline computed from average rewards of multiple sampled outputs per prompt, reducing memory overhead and boosting mathematical reasoning accuracy on GSM8K and MATH benchmarks. In listwise regimes, (5) Preference Ranking Optimization (PRO) \citep{song2024preference} formulates alignment as a Plackett-Luce ranking problem, optimizing the likelihood of full‐order permutations of candidate continuations to match human‐ranked preferences. (6) Listwise Preference Optimization (LiPO) \citep{liu2024lipo} formalizes this further through learning‐to‐rank objectives over graded response lists, leveraging the full spectrum of preference signals for more robust alignment. \textit{However, existing pairwise and listwise methods either reduce credit assignment to local comparisons or to static ranking of fixed candidate lists, and thus lack a mechanism to leverage full-trajectory exploration signals over long horizons. We overcome this gap in our proposed method by treating feedback on full outputs as scalar returns and and directly weighting entire candidate sequences by multiplying those scalar returns with the weights, yielding a low-variance, sample-efficient loss that naturally aligns model outputs with ranked preferences at the trajectory level, unlike token-wise methods which fragment the preference signal.}

\section{Proposed Method}
\label{sec:pl-model}

HAEPO adopts a \emph{trajectory-centric} view inspired by listwise ranking: every rollout
receives a single weight so that the entire history influences the update.

\subsection{Trajectory Log‑Likelihood}
\label{sec:traj-ll}

For each of the \(M\) complete episodes collected in a batch, we compute the cumulative
log‑likelihood
\begin{equation}
  L_k \;=\; \sum_{t=1}^{T_k} \log \pi_{\theta}\bigl(a^{(k)}_t \mid s^{(k)}_t\bigr),
  \qquad k=1,\dots,M,
\end{equation}
where,  \(T_k\) is the length of episode \(k\), \(s^{(k)}_t\) denotes the state encountered at time step \(t\) in episode \(k\), and \(a^{(k)}_t\) is the action taken by the policy in that state. 
Individual log‑probabilities are typically small (and negative). Hence, summing them
\emph{amplifies} the separation between trajectories, which are already favored by the current
policy and those it deems imperfect.

\subsection{Plackett-Luce Normalization}
\label{sec:pl-norm}
We transform these scores into listwise weights via the
Plackett-Luce softmax
\begin{equation}
  w_k \;=\; \frac{\exp(L_k)}{\sum_{j=1}^{M} \exp(L_j)}.
  \label{eq:pl-weights}
\end{equation}
Intuitively, \(w_k\) gives more credit to rollouts in which entire sequence of decisions is
consistent with the current policy, while down‑weighting unlikely or noisy episodes.
The exponential stretch provides the dynamic range needed to distinguish subtle
differences that would be lost with token‑level weighting.

\subsection{Listwise Policy Update}
\label{sec:listwise}
Using the weights \(w_k(\theta)\), we form a trajectory‑centric policy objective
that both rewards high‑return episodes and respects the policy’s own confidence.
By aggregating every time‑step into a single weight per rollout, sharper credit assignment and enhanced stability are achieved without discarding low‑probability
episodes entirely.

\subsection{Original Loss and Gradient}
\label{sec:orig-loss}

With the listwise weights \(w_k(\theta)\) defined in Eq.~\eqref{eq:pl-weights}, we now formalize the original HAEPO objective by rewarding each trajectory in proportion to its return under the current policy.

\paragraph{Original HAEPO loss.}
Rewarding each return in proportion to its weight gives the
\emph{original} HAEPO objective
\begin{equation}
  \mathcal{L}_{\mathrm{HAEPO}}^{\mathrm{orig}}(\theta)
    \;=\; - \sum_{k=1}^{M} R_k\,w_k(\theta).
  \label{eq:orig-plpo}
\end{equation}
where, \(R_k = \sum_{t=1}^{T_k} \gamma^{t-1} r^{(k)}_t\) denotes the cumulative discounted reward of episode \(k\) (with discount factor \(\gamma\)).  
(The minus sign converts the maximization of expected return into a minimization problem.)
\paragraph{Gradient.}
Starting from the definition of the trajectory weights,
\begin{equation}
  w_k \;=\; \frac{e^{L_k}}{\sum_{j=1}^M e^{L_j}}
  \quad\Longrightarrow\quad
  \log w_k = L_k - \log\Bigl(\sum_{j=1}^M e^{L_j}\Bigr),
\end{equation}
we obtain
\begin{equation}
  \nabla_\theta \log w_k
  = \nabla_\theta L_k
    - \frac{1}{\sum_{j}e^{L_j}}
      \sum_{j=1}^M e^{L_j}\,\nabla_\theta L_j \\[4pt]
  = \nabla_\theta L_k
    - \sum_{j=1}^M w_j\,\nabla_\theta L_j,
\end{equation}
i.e.\ a centered score function.  Now substitute into the score‐function gradient
\(\nabla_\theta\mathcal{L}_{\mathrm{HAEPO}}^{\mathrm{orig}}
 = -\sum_{k}R_k\,\nabla_\theta w_k\)
and use \(\nabla_\theta w_k = w_k\,\nabla_\theta\log w_k\) :
\begin{equation}
  \nabla_\theta \mathcal{L}_{\mathrm{HAEPO}}^{\mathrm{orig}}
  = -\sum_{k=1}^{M} R_k\,w_k\,\nabla_\theta \log w_k \\[4pt]
  = -\sum_{k=1}^{M} R_k\,w_k
     \Bigl[
       \nabla_\theta L_k
       - \sum_{j=1}^{M} w_j\,\nabla_\theta L_j
     \Bigr],
     \label{orig-haepo-grad}
\end{equation}

Each trajectory thus contributes its own score \(\nabla_\theta L_k\) offset by the
batch‑weighted average, so the policy \emph{leans in} toward episodes it both prefers
(high \(L_k\)) and deliver high return, while tempering updates for the rest.
This balanced, history‑aware gradient underpins HAEPO’s stable listwise credit assignment.

\subsection{Reward Normalization}
\label{sec:reward-norm}

To temper variance and ensure well‑scaled updates, we rescale the raw returns
\(\{R_k\}_{k=1}^M\) on every mini‑batch before applying the gradient in Eq.
\eqref{orig-haepo-grad}.  We considered two complementary schemes for the experiments:

\paragraph{Sum–Normalization.}
After optionally subtracting a baseline, we divide each return by the batch
sum,
\(
  \widetilde{R}_k = R_k\bigl/\sum_{j=1}^{M} R_j.
\)
This choice (1) constrains the rescaled rewards to a bounded value with
\(\sum_k \widetilde{R}_k = 1\), (2) emphasises \emph{relative} performance, which is an
advantage in one‑step or bandit settings where absolute scale is uninformative,
and (3) collapses to the classical Exp3/softmax update without bias when
\(T_k=1\).

\paragraph{Z–Score Normalization.}
For longer‑horizon tasks we instead centre and whiten the batch,
\(
  \widehat{R}_k = (R_k-\mu)/\sigma
\)
with
\(
  \mu = M^{-1}\sum_{j} R_j
\)
and
\(
  \sigma^2 = M^{-1}\sum_{j}(R_j-\mu)^2.
\)
This transformation (1) leaves the expectation unchanged so the gradient remains
unbiased, (2) provably lowers variance, which is critical in environments such as
\textsc{CartPole} where returns are highly dispersed, and (3) offers zero‑mean,
unit‑variance signals that pair well with adaptive optimisers like Adam,
thereby accelerating convergence.

In practice we select the scheme that best matches task horizon, but both integrate seamlessly into the
HAEPO gradient of Eq.~\eqref{orig-haepo-grad}, yielding a history‑aware update
with robust, stable credit assignment.

\subsection{Entropy-Regularised HAEPO}
\label{sec:entropy-plpo}

To prevent premature collapse onto a handful of fortunate trajectories, we add an
entropy bonus into the listwise objective.  Concretely, we maximise the
\emph{entropy-regularised} return
\begin{equation}
  \mathcal{L}_{\mathrm{ER}}(\theta)
  \;=\;
  -\sum_{k=1}^{M} w_k\,\widetilde{R}^{(\mathrm{norm})}_k
  \;+\;\beta\,\sum_{k=1}^{M} w_k \log w_k,
  \label{eq:er-loss}
\end{equation}
where, \(\beta \ge 0\) modulates the “keep-searching’’ pressure and
\(\widetilde{R}^{(\mathrm{norm})}_k\) denotes a reward rescaled by either
sum- or Z-score normalisation (Section~\ref{sec:reward-norm}).
The entropy term serves the purpose that when one rollout begins to dominate, its weight grows but the
\(\beta \sum_{k=1}^{M} w_k \log w_k\) penalty flattens the distribution, thus sustaining exploration.

\subsection{KL-Penalised HAEPO (Final Objective)}
\label{sec:kl-plpo}

A second stabiliser is a trajectory-level trust region that connects updates to
a reference policy \(\pi_{\mathrm{ref}}\).  For each episode we accumulate the
discounted divergence
\begin{equation}
  D_k
  \;=\;
  \log w_k(\theta)\;-\;\log w_k^{\mathrm{ref}},   
  \label{eq:traj-kl}
\end{equation}

and form the \emph{KL-penalised} objective
\begin{equation}
  \mathcal{L}(\theta)
  =\;
    -\sum_{k=1}^{M} w_k\,\widetilde{R}^{(\mathrm{norm})}_k
    \;+\;\beta\sum_{k=1}^{M} w_k \log w_k
  \\[6pt]
  \;+\;\lambda \sum_{k=1}^{M} w_k
          \bigl(\log w_k - \log w_k^{\mathrm{ref}}\bigr)\,,
\label{eq:final-loss}
\end{equation}
with \(\lambda \ge 0\) acting as a “speed governor’’ on policy shifts.
This trust-region term reins in high-variance gradients that could propel
\(\pi_{\theta}\) into low-density, poorly tested regions of the parameter space. 

Taken together, Eqs.~\eqref{eq:er-loss}-\eqref{eq:final-loss} define the
\textbf{final HAEPO loss}: a history-aware, entropy-tempered, KL-constrained
criterion that balances exploration, exploitation, and caution, which are crucial
ingredients for robust performance on both bandit and long-horizon control
tasks.

\subsection{Gradient Decomposition and Intuition}
\label{sec:grad-decomp}

The gradient of the full objective in Eq.~\eqref{eq:final-loss}
splits into three conceptually distinct forces:
\begin{equation}\label{eq:grad-split}
\nabla_\theta \mathcal{L}= -\,\mathbb{E}_{k\sim w(\theta)}
      \bigl[\tilde R_k^{(\mathrm{norm})}\,\nabla_\theta\log w_k\bigr]\\ +\,\beta\,
      \mathbb{E}_{k\sim w(\theta)}
      \bigl[\nabla_\theta\log w_k\bigr]\\ +\,\lambda\,
      \mathbb{E}_{k\sim w(\theta)}
      \bigl[D_k\,\nabla_\theta\log w_k\bigr].
\end{equation}
Reading from top to bottom:  
\emph{(1)~Reward seeking} steers the policy toward trajectories with large
normalised return, modulated by their current weight;  
\emph{(2)~Entropy pressure} spreads the weights, forestalling early
concentration;  
\emph{(3)~Weight-space KL} acts as a batch-level trust region, damping abrupt
re-weighting.  
Collectively, these terms negotiate exploration, exploitation, and stability.

For brevity, Eq.~\eqref{eq:grad-split} can be collapsed to
\begin{equation}
\label{eq:grad-compact}
\nabla_\theta \mathcal{L}
= -\,\mathbb{E}_{k\sim w(\theta)}
    \Bigl[\bigl(\tilde R_k^{(\mathrm{norm})} - \beta - \lambda D_k\bigr)
          \,\nabla_\theta \log w_k\Bigr].
\end{equation}
This centred score-function form highlights why the updated HAEPO remains
low-variance even with the weight-level KL regulariser.

\section{Evaluation}
We evaluated HAEPO in four different environments to highlight its key properties and novelties:

\subsection{Multi-armed Bandit}

The goal in a stochastic multi-armed bandit \citep{robbins1952some,Auer2002FinitetimeAO} is to minimize per-step regret. It is defined as the gap at each pull between the reward actually received and the reward that would have received by always pulling the optimal arm, while still exploring enough to discover which arm is best.  HAEPO’s listwise Plackett-Luce weighting naturally balances this trade-off by amplifying high-return arms. Also, via an entropy bonus, HAEPO keeps weaker arms in play long enough to reduce premature exploitation.

\paragraph{Setup \& Baselines.}
We test on stochastic Gaussian bandits with $K$ arms, varying $K\in\{10,20,30\}$.  Each arm $k$ yields rewards $r\sim\mathcal{N}(\mu_k,1)$, where $\mu_k$ is drawn uniformly from $[0,1]$ and held fixed.  For each $(K,\mathrm{algorithm})$ pair we sweep:

\[
\begin{aligned}
\text{learning rates} &\in \{1\times10^{-3},\,2\times10^{-3},\,5\times10^{-3}\},\\
\text{batch sizes}  &\in \{8,16,32\}.
\end{aligned}
\]
and average results over $5$ random seeds.  We compare:
\begin{itemize}
  \item \textbf{HAEPO}, with entropy weight $\beta_{\rm ent}=5\times10^{-2}$ and KL penalty $\beta_{\rm kl}=5\times10^{-2}$.
  \item \textbf{PPO}-style policy gradient (per-step updates, no trajectory normalization) clipped-surrogate ($\epsilon$ = 0.2).
  \item \textbf{DPO} (trajectory-level binary preference optimization).
\end{itemize}

\paragraph{Metrics \& Evaluation.}
After $T=5{,}000$ pulls we record \textit{per-step regret} (mean $\pm$ standard deviation over seeds).

\begin{figure}[H]
  \centering
  \includegraphics[width=\textwidth]{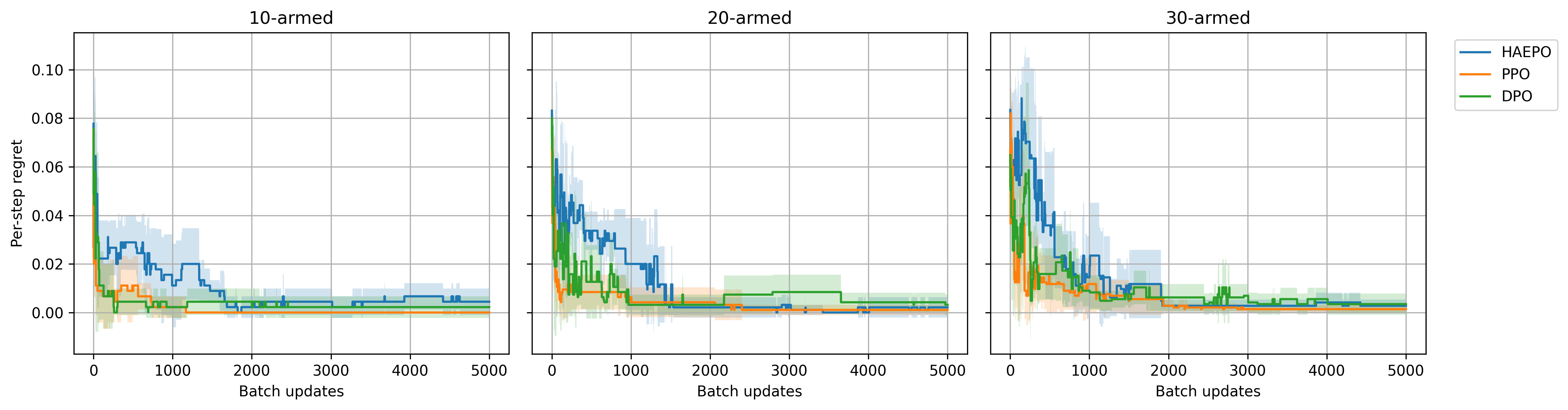}
  \caption{Multi-armed Bandit Environment: Mean per-step regret over $5{,}000$ pulls for $K\in\{10,20,30\}$.}
  \label{fig:exp-bandit}
\end{figure}

\paragraph{Results.} 
As shown in Fig.~\ref{fig:exp-bandit}, we conclude the following: 
\begin{itemize}
  \item \emph{Increased exploration with more arms:}  
    As $K$ grows, HAEPO’s policy entropy curves rise, showing that it sustains exploration longer when faced with more options as compared to PPO and DPO.
  \item \emph{Gained stability:}  
    Fluctuations in HAEPO’s per-step regret shrink for larger $K$, indicating that it becomes more stable as the number of arms increases as compared to PPO and DPO.
  \item \emph{Fast convergence:}  
    Across all $K$, HAEPO reaches low per-step regret at a similar speed to PPO and DPO, without sacrificing exploration or stability.
\end{itemize}

These observations highlight that HAEPO not only explores more effectively as task complexity scales, but also gains robustness, hence staying both exploratory and stable while converging as quickly as standard baselines.

\subsection{Random Walk}

Sparse, long-horizon tasks, such as a random walk environment \citep{sutton2018reinforcement}, are notoriously challenging for per-step policy gradients. In such environments, a single nonzero reward only at the end yields vanishingly small gradient signals until many samples are collected. HAEPO compresses each entire trajectory of length $T=500$ into one log-likelihood score, then applies listwise Plackett-Luce weighting to sharply amplify rare successful walks.

\paragraph{Setup \& Baselines.}
We use a one-dimensional random-walk environment of fixed length $T=500$. The agent starts at position 0 and chooses steps $a_t\in\{\pm1\}$ until it reaches $+T$ (success) or exhausts all steps (failure). Each trajectory yields a return $R\in\{0,1\}$. We compare:
\begin{itemize}
  \item \textbf{HAEPO}, with $\beta_{\rm ent}=5\times10^{-5}$ and $\beta_{\rm kl}=5\times10^{-5}$.
  \item \textbf{PPO}, clipped-surrogate ($\varepsilon=0.2$).
  \item \textbf{DPO}, trajectory-level binary preference optimization.
\end{itemize}
Each algorithm runs for $n_{\mathrm{updates}}=100$ gradient steps with batch size 32 (3,200 trajectories per seed) over 5 seeds.

\begin{figure}[H]
  \centering
  \includegraphics[width=0.85\textwidth]{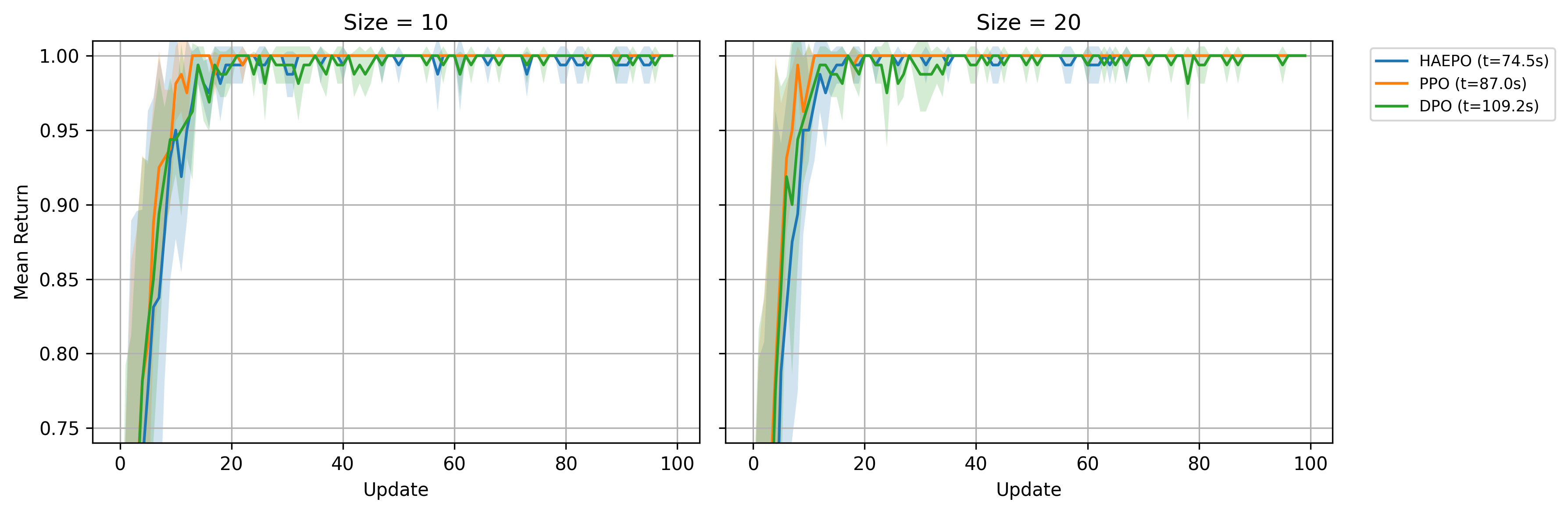}
  \caption{Random Walk Environment: Mean return over 100 updates for state sizes 10 (left) and 20 (right). Legend indicates average wall-clock time per 100 updates.}
  \label{fig:exp-rwalk}
\end{figure}

\paragraph{Metrics \& Evaluation.}
We track the \emph{mean return} (fraction of successful walks) as a function of gradient updates. Figure~\ref{fig:exp-rwalk} plots mean return over 100 updates for state sizes $n=10$ (left) and $n=20$ (right). We also measure \textit{average wall-clock time} per 100 updates. Legend entries report average wall-clock time per 100 updates.

\paragraph{Results.} As shown in Fig.~\ref{fig:exp-rwalk}, we can deduce the following advantages of HAEPO relative to PPO and DPO : 
\begin{itemize}
  \item \emph{Fast convergence:} HAEPO reaches near-optimal mean return within about 12 updates, matching PPO and DPO in speed.
  \item \emph{Improved stability with scale:} Variance in HAEPO’s return curve decreases when moving from size 10 to size 20, showing that it gains robustness as the state space grows.
  \item \emph{Compute efficiency:} HAEPO runs $\sim15\%$ faster than PPO and $\sim32\%$ faster than DPO per 100 updates.
  \item \emph{Strong exploration:} Trajectory-level weighting preserves exploration early in training, sustaining diversity without sacrificing convergence for HAEPO as compared to PPO and DPO.
\end{itemize}

These findings confirm that HAEPO maintains fast learning, solid exploration, and growing stability even as task complexity increases, thus outperforming standard per-step and PPO-style baselines in sparse, long-horizon settings.

\subsection{CartPole (Gymnasium)}  
CartPole \citep{6313077,1606.01540} is a classic control benchmark where the agent must balance a pole on a cart by choosing left/right forces. Training uses a two‐layer MLP (obs → 128 → 2, ReLU), with learning rate \(\alpha=10^{-2}\), discount \(\gamma=0.99\), batch size \(M=8\), and up to 500 updates.

\paragraph{Setup \& Baselines.}
We use Gymnasium’s \texttt{CartPole-v1} environment. Each algorithm is run for up to 200 s of wall‐clock time, over five random seeds \(\{0,1,2,3,4\}\). We compare:
\begin{itemize}
  \item \textbf{HAEPO}: entropy bonus \(\beta_{\rm ent}=0.1\), KL penalty \(\beta_{\rm kl}=0.1\), gradient‐norm clipping \(\max\|\nabla\|=0.5\).
  \item \textbf{PPO}: clipped surrogate with \(\varepsilon=0.2\), one PPO epoch per update, no explicit entropy bonus.
  \item \textbf{HAEPO w/o reg.}: same as HAEPO but with \(\beta_{\rm ent}=0\) and \(\beta_{\rm kl}=0\).
\end{itemize}

\paragraph{Wall‐clock learning curves.}
Figure~4 plots \textit{the mean return versus time} (mean±std) aggregated across the five seeds (interpolated to a common \qty{0}-\qty{200}{\second} grid). PPO reaches the max-return in roughly \qty{40}{\second}, while HAEPO attains the same level in about \qty{130}{\second}. The unregularized variant converges much slower (approx. \qty{140}{\second}) and exhibits large oscillations.

\begin{figure}[H]
  \centering
  \includegraphics[width=0.65\textwidth]{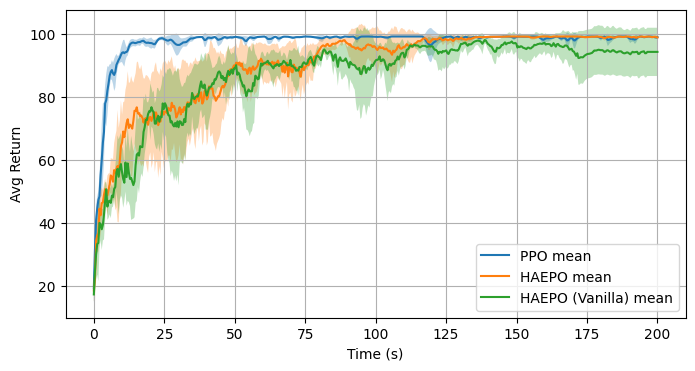}
  \caption{Wall-clock learning curves on the CartPole-v1 benchmark, showing the average return (mean ± 1 std across five seeds) as a function of elapsed time.}
  \label{fig:exp-cartpole}
\end{figure}

\paragraph{Results.} From Fig.~\ref{fig:exp-cartpole}, we conclude the following points:
\begin{itemize}
  \item \emph{Slower but smoother convergence:}  
    PPO solves in \(\approx\qty{40}{\second}\), whereas HAEPO requires \(\approx\qty{130}{\second}\), about 3× slower, but its learning curve is markedly smoother.
  \item \emph{Stable performance plateau:}  
    Once HAEPO achieves the threshold, its curve remains tightly clustered with negligible drift.
  \item \emph{Regularization drives stability:}  
    The unregularized variant oscillates heavily and only solves after \(\approx\qty{140}{\second}\), underscoring that both entropy and KL penalties are critical for HAEPO’s consistency.
\end{itemize}

These findings demonstrate that HAEPO not only produces competitive sample efficiency on CartPole but, more importantly, delivers dramatically more stable and predictable policies, reducing uncertainty in deployment scenarios.

\subsection{LLM TL;DR} 

The TL;DR \citep{volske-etal-2017-tl} task challenges models to distill lengthy Reddit posts into concise, informative summaries under a reinforcement‐learning fine‐tuning regime. By pairing 1000 supervised examples with 1000 RL updates per seed (batch size = 4), we evaluate summary quality (ROUGE-L, Semantic Similarity, human preference), style (brevity, profanity incidence, response length), and efficiency (peak VRAM).


\begin{table}[H]
  \centering
  \small                    
  \setlength{\tabcolsep}{6pt}
  \caption{Human evaluation (mean $\pm$ std) on the TL;DR dataset (5 seeds).}
  \label{tab:human-eval-tldr}
  \begin{tabular}{@{}lccc@{}}
    \toprule
    Model & HAEPO & DPO & GRPO \\
    \midrule
    Llama\,3.2 (1\,B)  & \textbf{4.4\,$\pm$\,0.7} & 3.5\,$\pm$\,0.2 & 3.9\,$\pm$\,0.8 \\
    Qwen\,2.5 (1.5\,B) & \textbf{4.3\,$\pm$\,0.2} & 3.7\,$\pm$\,1.2 & 4.0\,$\pm$\,1.2 \\
    \bottomrule
  \end{tabular}
\end{table}

\paragraph{Setup \& Baselines.}
We fine-tune two model variants: LLaMA 3.2 (1B) \citep{dubey2024llama,meta_llama3_2_2024} and Qwen 2.5 (1.5B) \citep{yang2025qwen3}, on the TL;DR dataset using a 50-50 split of 1,000 supervised examples and 1,000 RL updates per seed (batch size = 4; RL learning rate = $5 \times 10^{-5}$). Each method starts from the same checkpoint and is run with three random seeds. The training was done on A100 GPUs (40GB). We compare the following : 
\begin{itemize}
  \item \textbf{HAEPO (\(\beta_{\rm ent}= 1\times 10^{-2}\), \(\beta_{\rm kl}=1\times 10^{-2}\)).} Reward model combines F1, semantic similarity, and length closeness.
  \item \textbf{GRPO.} Token-level listwise ranking with identical entropy/KL settings and reward model same as HAEPO.
  \item \textbf{DPO.} Pairwise Direct Preference Optimization (gold-vs-model reference policy).
\end{itemize}

\paragraph{Metrics \& Evaluation.}
\begin{itemize}
  \item \emph{End‐to‐end training time:} wall‐clock minutes for the full fine‐tuning pipeline (data loading, supervision + RL, and evaluation).  
  \item \emph{Peak VRAM usage:} maximum GPU memory consumed during each run.
  \item \emph{Alignment rate:} fraction of human A/B votes in which the preferred summary matches the original poster’s TL;DR, indicating alignment with the author’s intent.
\end{itemize}

\paragraph{Results.}
Table~\ref{tab:human-eval-tldr} reports the mean human preference rating (1–5 scale) ± 1 std across 40 participants evaluating summaries generated from five random seeds.  For LLaMA 3.2, HAEPO achieves a rating of \(\mathbf{4.4\pm0.7}\), outperforming GRPO (\(3.9\pm0.8\)) and DPO (\(3.5\pm0.2\)).  For Qwen 2.5, HAEPO likewise leads with \(4.3\pm0.2\), versus GRPO’s \(4.0\pm1.2\) and DPO’s \(3.7\pm1.2\).  Not only does HAEPO secure the highest average preference, but it also exhibits lower variance indicating more consistent alignment with human judgments.  
Notably, HAEPO trained faster than GRPO and DPO completing in just 20 minutes compared to 25 and 28 minutes respectively on Qwen 2.5 (1.5B). In terms of GPU usage, HAEPO consumes 28GB peak VRAM, less than GRPO (38GB) but slightly more than DPO (26GB), highlighting a favorable efficiency performance trade-off on the same model.

\paragraph{Qualitative Feedback.}
Open-ended comments show that DPO’s richer, “dramatic” summaries were praised for detail but felt too long, while GRPO’s occasional profanity was jarring. HAEPO consistently delivered concise, profanity-free summaries and achieved the highest human preference rates.

\section{Limitations}

We note the following limitations of HAEPO : 

\begin{itemize}
  \item \textbf{Hyper‐parameter sensitivity:}  
    Although HAEPO uses only two main regularization scalars (\(\beta_{\rm ent},\beta_{\rm kl}\)), their optimal settings can vary substantially across tasks and reward scales.  Future work should explore automated tuning, adaptive schedules (e.g.\ meta‐gradient, population‐based), or even dynamic schedulers (e.g.\ annealing or cosine‐decay) for \(\beta_{\rm ent}\) and \(\beta_{\rm kl}\) to further improve convergence. 
  \item \textbf{Extremely long horizons:}  
    We demonstrate HAEPO up to \(\sim10^3\) time-steps, but real-world domains (e.g.\ Minecraft or StarCraft II) can span \(10^4\!+\) steps.  Handling such horizons may require incorporating memory-efficient episode buffering, truncated backpropagation, or hierarchical decomposition in HAEPO.  
  \item \textbf{Computational overhead:}  
    The listwise softmax over \(M\) trajectories adds \(O(M)\) extra work each update and requires storing per-episode log-prob sums.  In settings with very large batch sizes or resource-constrained devices, this overhead may be non-negligible. A similar storage overhead applies to other listwise/ranking-based methods (e.g.\ GRPO), while pairwise methods (DPO) and per-step methods (PPO) incur their own, but differently structured memory and computation costs.
  \item \textbf{Single‐agent focus:}  
    All experiments are in single‐agent environments.  Multi-agent or competitive settings introduce new challenges (e.g.\ non-stationarity, joint trajectory ranking) that HAEPO does not yet address.
  \item \textbf{Lack of large-scale LLM benchmarks:}  
    We did not include AlpacaEval \citep{alpaca_eval} or MT-Bench \citep{bai2024mt} in our experiments due to the absence of standardized public datasets and limited GPU VRAM, which prevented large-model inference and evaluation. To partially compensate, we used the TL;DR \citep{volske-etal-2017-tl} dataset as a lightweight with human preferences using forms. Future work could incorporate these benchmarks given access to sufficient computational resources.

\end{itemize}

\section{Conclusion}

We introduced \textbf{History‑Aggregated Exploratory Policy Optimization (HAEPO)}, a trajectory‑level policy‑gradient method that couples Plackett-Luce weighting with entropy and soft‑KL regularisation. By collapsing every episode into a single log‑likelihood score, HAEPO captures long‑range credit assignment signals that token‑ and step‑wise objectives dilute, while its dual regularisers maintain both exploration and stability. Empirically, HAEPO converges fast, explores thoroughly, aligns closely with true rewards, and demonstrates robust learning behavior better or at par with PPO, GRPO, and DPO across diverse tasks. Thus, HAEPO provides a stable and interpretable framework by explicitly leveraging full‑trajectory history while balancing exploration and stability. As part of the future work, we plan to overcome the limitations mentioned in the previous section.

\bibliographystyle{plainnat}
\bibliography{citation}

\appendix

\section*{Comprehensive Mathematical Derivation of HAEPO Loss}
\label{appendix:haepo-full}

This appendix provides an exhaustive derivation of the HAEPO loss and its gradient, including proofs of key properties, variance reduction analysis, and detailed sign justification for regularization terms.
All objectives below are written for \textit{minimisation}. Consequently, maximising expected return appears with a leading minus sign in $\mathcal{L}^{\mathrm{orig}}$ and its descendants.

\subsection*{Motivation for HAEPO Loss}
Traditional policy-gradient methods predominantly operate at the per‐step level, treating each action independently and thereby diluting the credit assignment signal over long trajectories.  This fragmentation often leads to high variance updates and premature convergence to suboptimal policies.  Meanwhile, trust‐region approaches like PPO enforce stability but do not directly address the loss of global trajectory structure, and entropy bonuses are typically applied at the action level rather than holistically.  In parallel, listwise ranking models in information retrieval, specifically the Plackett-Luce model’s first‐choice probability, demonstrated the power of weighting entire lists based on a single score.

HAEPO arises by unifying three complementary strands of prior work that, in isolation, fell short of tackling long‐horizon, low‐variance policy learning:
\begin{itemize}
  \item \textbf{Trajectory‐centric credit assignment:} Unlike per‐step score functions that fragment credit and inflate variance, listwise ranking models (e.g.\ Plackett-Luce first‐choice probabilities) naturally weight entire sequences based on a single score.
  \item \textbf{Exploration via entropy:} Classic entropy bonuses act locally on action probabilities, but applying entropy directly to a trajectory‐weight distribution preserves global rollout diversity.
  \item \textbf{Stability via trust regions:} PPO‐style KL penalties constrain incremental policy shifts, yet they do not exploit trajectory structure or reduce variance via batch‐level weighting.
\end{itemize}
By taking the first‐choice probability from the Plackett-Luce model over full rollouts, HAEPO aggregates return signals into one coherent weight per trajectory.  Layering an entropy regularizer on this weight distribution then ensures persistent exploration, while a KL trust‐region penalty anchors updates to a reference policy for stability.  In this way, HAEPO “connects the dots” between listwise ranking theory, global entropy incentives, and trust‐region constraints, yielding a single, unbiased, low‐variance update rule tailored for long‐horizon reinforcement learning.

\subsection*{Notation and Preliminaries}
\subsubsection*{Trajectory Batch}
Let $\mathcal{B} = \{\tau_k\}_{k=1}^M$ be a batch of $M$ trajectories (or episodes).  For each $k$, let
\begin{equation}
\tau_k = \bigl(s^{(k)}_1,\,a^{(k)}_1,\,r^{(k)}_1,\;\dots,\;s^{(k)}_{T_k},\,a^{(k)}_{T_k},\,r^{(k)}_{T_k}\bigr)
\end{equation}
denote the $k$-th trajectory of length $T_k$.  Here, $s^{(k)}_t$ is the state at time $t$, $a^{(k)}_t$ is the action taken, and $r^{(k)}_t$ is the corresponding reward.

\subsubsection*{Log-likelihood \& Return}
For each trajectory $\tau_k$, define:
\begin{equation}
L_k(\theta) = \sum_{t=1}^{T_k} \log \pi_\theta\bigl(a^{(k)}_t \mid s^{(k)}_t\bigr),
\end{equation}
\begin{equation}
R_k = \sum_{t=1}^{T_k} \gamma^{t-1} r^{(k)}_t,
\end{equation}
where $\pi_\theta$ is the parameterized policy and $\gamma\in[0,1]$ the discount factor.

\subsection*{Plackett--Luce Weights}
The Plackett--Luce (PL) weight \citep{plackett1975analysis} for trajectory $\tau_k$ is
\begin{equation}
w_k(\theta) = \frac{\exp(L_k(\theta))}{\sum_{j=1}^M \exp(L_j(\theta))},
\end{equation}
\begin{equation}
\sum_{k=1}^M w_k(\theta) = 1.
\end{equation}
\textit{Moreover, this weight $w_k(\theta)$ can be interpreted as the marginal probability of selecting trajectory $k$ first under the Plackett-Luce model over the $M$ trajectories. That is, it is not merely a softmax but the first-choice probability in the PL distribution.}

\subsection*{Differentiation of PL Weights}
Using $f(L)=e^L$ and the quotient rule, we derive:
\begin{equation}
\nabla_\theta w_k = \frac{e^{L_k}\nabla_\theta L_k\, (\sum_j e^{L_j}) - e^{L_k} (\sum_j e^{L_j}\nabla_\theta L_j)}{(\sum_j e^{L_j})^2}
= w_k \bigl(\nabla_\theta L_k - \sum_{j=1}^M w_j\nabla_\theta L_j\bigr).
\end{equation}
Define the \emph{score-function} form:
\begin{equation}
\nabla_\theta \log w_k
= \nabla_\theta L_k - \sum_{j=1}^M w_j(\theta)\,\nabla_\theta L_j.
\label{eq:score}
\end{equation}

\subsection*{Original HAEPO Loss}
The base HAEPO objective to minimize is
\begin{equation}
\mathcal{L}^{\mathrm{orig}}(\theta)
= - \sum_{k=1}^M R_k \, w_k(\theta).
\end{equation}

\subsection*{Gradient of Original Loss}
\label{prop:orig_grad}
Its gradient is
\begin{equation}
\nabla_\theta \mathcal{L}^{\mathrm{orig}}
= -\sum_{k=1}^M R_k \, w_k \, \nabla_\theta \log w_k,
\end{equation}
which, by substituting \eqref{eq:score}, yields
\begin{equation}
\nabla_\theta \mathcal{L}^{\mathrm{orig}}
= -\sum_{k=1}^M R_k\,w_k
  \Bigl(\nabla_\theta L_k - \sum_{j=1}^M w_j\nabla_\theta L_j\Bigr).
\end{equation}

Using $\nabla w_k = w_k\nabla\log w_k$ and linearity of gradients.

\subsection*{Reward Normalization}
To further control scale of $R_k$, one may define normalized returns:
\begin{equation}
\widetilde R_k = \frac{R_k}{\sum_j R_j},
\quad
\widehat R_k = \frac{R_k - \mu_R}{\sigma_R},
\end{equation}
with sample mean $\mu_R$ and std $\sigma_R$. Both preserve $\mathbb{E}[\nabla\mathcal{L}]$.

\subsection*{Entropy Regularization}
\subsubsection*{Entropy Bonus}
The entropy of $w$ is
\begin{equation}
H(w) = -\sum_{k=1}^M w_k \log w_k.
\end{equation}
To encourage exploration, we add $-\beta H(w)$ (with $\beta>0$), yielding:
\begin{equation}
\mathcal{L}^{\rm ent}(\theta)
= \mathcal{L}^{\rm orig}(\theta) - \beta H(w)
= -\sum_k R_k w_k + \beta \sum_k w_k\log w_k.
\end{equation}

\subsubsection*{Sign interpretation}
Recall that $w_k\log w_k\le 0$ for every $k$, so the additive term
$\beta\sum_{k} w_k\log w_k$ \emph{decreases} the objective whenever the
entropy grows.  Thus, choosing $\beta>0$ correctly rewards broader, more
exploratory weight distributions.

\subsection*{KL Trust-Region Penalty}
\subsubsection*{KL Divergence}
Between current and reference PL weights $w^{\rm ref}$,
\begin{equation}
\mathrm{KL}(w\|w^{\rm ref})
= \sum_k w_k(\log w_k - \log w_k^{\rm ref}).
\end{equation}
where the reference weights are computed at frozen policy parameter \(\theta_{\rm ref}\), e.g. the previous policy update:
\begin{equation}
w_k^{\rm ref}
= \frac{\exp\!\bigl(L_k(\theta_{\rm ref})\bigr)}
       {\sum_{j=1}^M \exp\!\bigl(L_j(\theta_{\rm ref})\bigr)},
\end{equation}
\begin{equation}
w^{\rm ref} = \bigl(w_1^{\rm ref}, \dots, w_M^{\rm ref}\bigr).
\end{equation}

Including $+\lambda\,\mathrm{KL}(w\|w^{\rm ref})$ gives the final loss:
\begin{equation}
\mathcal{L}(\theta)
= \mathcal{L}^{\rm ent}(\theta) + \lambda\,\mathrm{KL}(w\|w^{\rm ref}).
\end{equation}

\subsubsection*{Constant-term cancellation in the KL gradient}
Differentiating $\mathrm{KL}(w\|w^{\mathrm{ref}})=\sum_k
w_k(\log w_k-\log w_k^{\mathrm{ref}})$ yields a factor
$1+\log w_k-\log w_k^{\mathrm{ref}}$.  
The constant $+1$ vanishes because
\begin{equation}
\textstyle\sum_{k} w_k\nabla_\theta\log w_k
      = \sum_{k}\nabla_\theta w_k
      = \nabla_\theta\!\bigl(\sum_k w_k\bigr)
      = 0,
\end{equation}
hence we drop it for notational economy without affecting the result.

\subsection*{Final Gradient Form}
Combining all original and regularization terms, the gradient is
\begin{equation}
\nabla_\theta \mathcal{L}
= -\sum_{k=1}^M w_k \bigl[\,R_k
   - \beta\,(1+\log w_k)
   - \lambda\,D_k \bigr] 
\;\times \bigl(\nabla_\theta L_k
   - \sum_{j=1}^M w_j\,\nabla_\theta L_j \bigr).
\end{equation}
where \(D_k=\log w_k-\log w_k^{\rm ref}\).  Equivalently, in expectation form:
\begin{equation}
\nabla_\theta \mathcal{L}
= -\mathbb{E}_{k\sim w}\Bigl[\bigl(R_k 
   - \beta(1+\log w_k) 
   - \lambda\,D_k\bigr)
\;\times \nabla_\theta \log w_k\Bigr].
\end{equation}

Or broken out into separate score‐function components:
\begin{equation}
\nabla_\theta \mathcal{L}
= -\mathbb{E}_{k\sim w}\bigl[R_k\,\nabla_\theta\log w_k\bigr]
   \;+ \beta\,\mathbb{E}_{k\sim w}\bigl[(1+\log w_k)\,\nabla_\theta\log w_k\bigr]
\;+ \lambda\,\mathbb{E}_{k\sim w}\bigl[D_k\,\nabla_\theta\log w_k\bigr].
\end{equation}

\subsection*{Unbiasedness of the HAEPO Gradient Estimator}
\label{sec:unbiased}
We now give a complete proof that our Monte Carlo estimator of the HAEPO gradient is unbiased.  Crucially, we must account not only for the \(\theta\)-dependence of the trajectory weights \(w_k\), but also for the \(\theta\)-dependence of the sampling distribution \(p(\tau;\theta)\).  We proceed via continuous integrals, the score‐function (log‐derivative) trick, and iterated expectations.

\subsubsection*{Sampling and Joint Density.}
Let \(\tau=(\tau_1,\dots,\tau_M)\) be \(M\) independent and identically distributed full rollouts drawn from policy \(\pi_\theta\).  Their joint density factors as
\begin{equation}
  p(\tau;\theta)
  \;=\;
  \prod_{i=1}^M p(\tau_i;\theta)
  \;=\;
  \prod_{i=1}^M \pi_\theta(\tau_i).
\end{equation}

\subsubsection*{True Gradient as an Integral.}
Define the loss
\begin{equation}
  \mathcal{L}(\theta)
  = -\,\mathbb{E}_{p(\tau;\theta)}\Bigl[\sum_{k=1}^M R_k(\tau_k)\,w_k(\tau;\theta)\Bigr].
\end{equation}
Then by the product‐rule for \(\nabla_\theta\),
\begin{equation}
\begin{split}
G(\theta)
&=-\nabla_\theta \int \Bigl(\sum_{k=1}^M R_k\,w_k\Bigr)\,p(\tau;\theta)\,d\tau\\
&=-\int \nabla_\theta\Bigl(\sum_{k=1}^M R_k\,w_k\Bigr)\,p\,d\tau -\;\int \Bigl(\sum_{k=1}^M R_k\,w_k\Bigr)\,\nabla_\theta p\,d\tau\\
&=-\int \sum_{k=1}^M R_k\,w_k\,\nabla_\theta\log w_k\;p\,d\tau -\;\int \Bigl(\sum_{k=1}^M R_k\,w_k\Bigr)\,\nabla_\theta\log p\;p\,d\tau.
\end{split}
\label{eq:true_grad_corrected}
\end{equation}
Here we used \(\nabla_\theta w_k = w_k\,\nabla_\theta\log w_k\) and \(\nabla_\theta p = p\,\nabla_\theta\log p\).
(using the score‐function identity; see \citep{Glynn1990,williams1992simple}).

\subsubsection*{Monte Carlo Estimator}
On a single batch \(\tau\), an unbiased estimator that matches both integrals in \eqref{eq:true_grad_corrected} is
\begin{equation}
\hat G_M(\theta)
= -\sum_{k=1}^M R_k(\tau_k)\,w_k(\tau;\theta)\,\nabla_\theta\log w_k(\tau;\theta)
\;-\;\Bigl(\sum_{k=1}^M R_k(\tau_k)\,w_k(\tau;\theta)\Bigr)\,\nabla_\theta\log p(\tau;\theta).
\end{equation}
(cf.\ the policy‐gradient theorem; see \citep{sutton2018reinforcement})

\subsubsection*{Unbiasedness via the Score‐Function Trick}
By the definition of expectation,
\begin{equation}
\mathbb{E}[\hat G_M(\theta)]
=-\int \sum_{k=1}^M R_k\,w_k\,\nabla_\theta\log w_k\;p\,d\tau -\int \Bigl(\sum_{k=1}^M R_k\,w_k\Bigr)\,\nabla_\theta\log p\;p\,d\tau=G(\theta),
\end{equation}
so \(\mathbb{E}[\hat G_M(\theta)]=G(\theta)\), establishing unbiasedness.

\subsubsection*{Law of Iterated Expectation}
Equivalently, one may condition on \(\tau_1,\dots,\tau_{M-1}\) and then on \(\tau_M\), iteratively, to recover the same two terms and conclude \(\mathbb{E}[\hat G_M]=G(\theta)\).

\subsubsection*{Variance Characterization}
The variance of this corrected estimator is
\begin{equation}
\mathrm{Var}[\hat G_M]
=\mathbb{E}\bigl[\|\hat G_M\|^2\bigr] - \|G(\theta)\|^2,
\end{equation}
which now reflects contributions from both the weight‐gradient term and the sampling‐distribution term.

\subsubsection*{Extension to Regularization}
Exactly the same argument applies to the entropy and KL‐penalty components once you include their dependence on \(\theta\) both through \(w_k\) and through \(p(\tau;\theta)\); each score‐function integral picks up two pieces and their Monte Carlo estimators remain unbiased.

\section*{Code Sanity Check}
\subsection*{Gradient Verification}
To verify our analytic gradients for the HAEPO loss, we applied a central finite‐difference check with perturbation $\epsilon = 10^{-6}$. For each input : newlp, oldlp, and returns, we compared
\[
  \frac{\partial \mathcal{L}}{\partial x_i}
  \quad(\text{autodiff})
  \quad\text{against}\quad
  \frac{\mathcal{L}(x_i + \epsilon) - \mathcal{L}(x_i - \epsilon)}{2\epsilon}
  \quad(\text{numeric}).
\]
The maximum absolute difference was on the order of $10^{-11}$, well below our $10^{-6}$ threshold, demonstrating consistency in our gradient computations.

\paragraph{Results.}
\begin{itemize}
  \item \emph{Maximum absolute gradient errors:}
    \begin{itemize}
      \item new\_lp  : \(5.50\times 10^{-11}\)
      \item old\_lp  : \(1.62\times 10^{-11}\)
      \item returns  : \(8.31\times 10^{-11}\)
    \end{itemize}
  \item \emph{Gradient norms:}
    \begin{itemize}
      \item \(\|\nabla_{\text{new\_lp}}\| = 0.438904\)
      \item \(\|\nabla_{\text{old\_lp}}\| = 0.004911\)
      \item \(\|\nabla_{\text{returns}}\| = 0.631110\)
    \end{itemize}
\end{itemize}

\subsection*{Choosing Reward Normalization}
To illustrate when to use sum-normalization vs.\ z-score normalization in HAEPO, we conduct two lightweight ablations on tasks with contrasting reward structures.

\subsubsection*{When to choose which norm}
\begin{itemize}
  \item \textbf{Sum-Normalization}  
    Ideal for one-step or low-variance, dense‐reward tasks (e.g.\ single-period newsvendor).  
    \begin{itemize}
      \item Preserves absolute reward scale ($\sum_k \tilde R_k = 1$).  
      \item Emphasises relative performance when returns lie in a narrow band.  
      \item Collapses to Exp3/softmax update for $T_k=1$ without bias.
    \end{itemize}
  \item \textbf{Z-Score Normalization}  
    Ideal for long-horizon or high‐variance, sparse-reward tasks (e.g.\ deterministic chain MDP).  
    \begin{itemize}
      \item Centres and whitens returns, giving zero-mean, unit-variance signals.  
      \item Keeps the gradient estimator unbiased while provably reducing variance.  
      \item Pairs well with adaptive optimisers (e.g.\ Adam), accelerating convergence.
    \end{itemize}
\end{itemize}

\begin{figure}[H]
  \includegraphics[width=\textwidth]{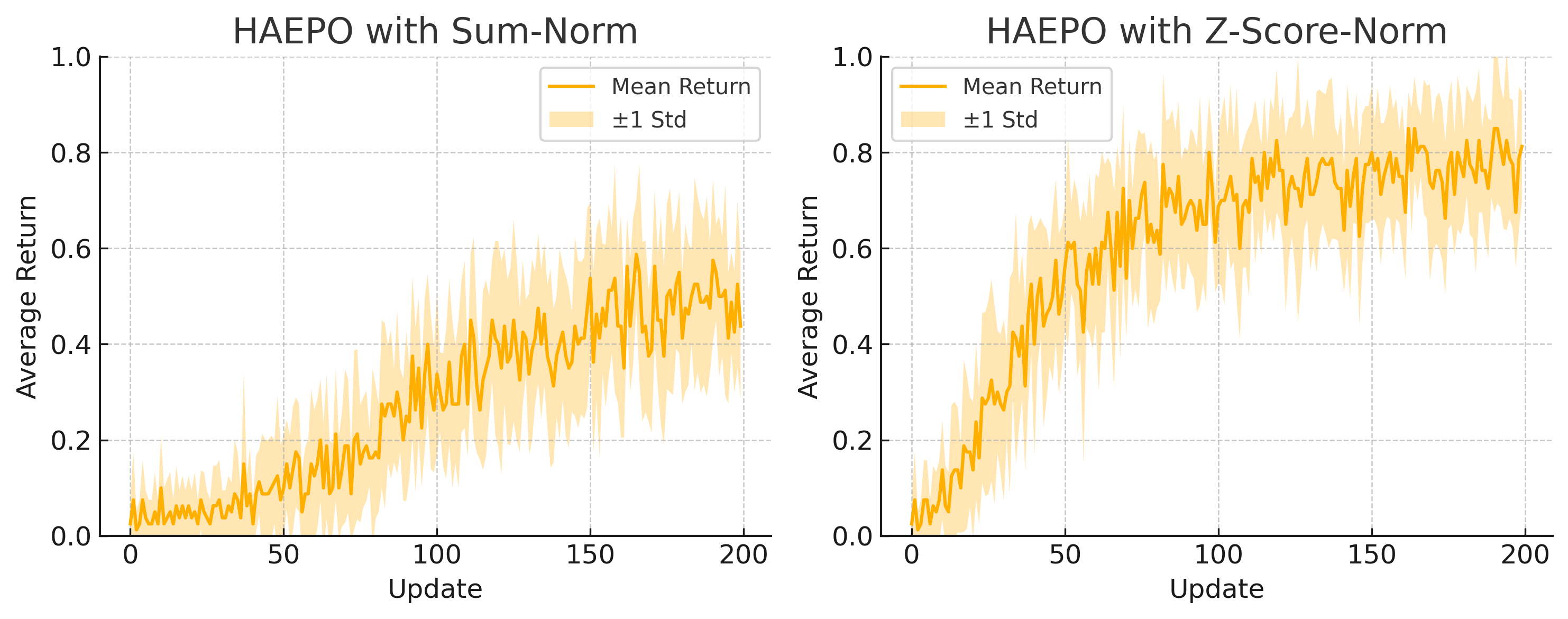}
  \caption{HAEPO on the Deterministic Chain MDP: sum-norm vs.\ z-score.}
  \label{fig:chain-mdp}
\end{figure}

\begin{figure}[H]
  \includegraphics[width=\textwidth]{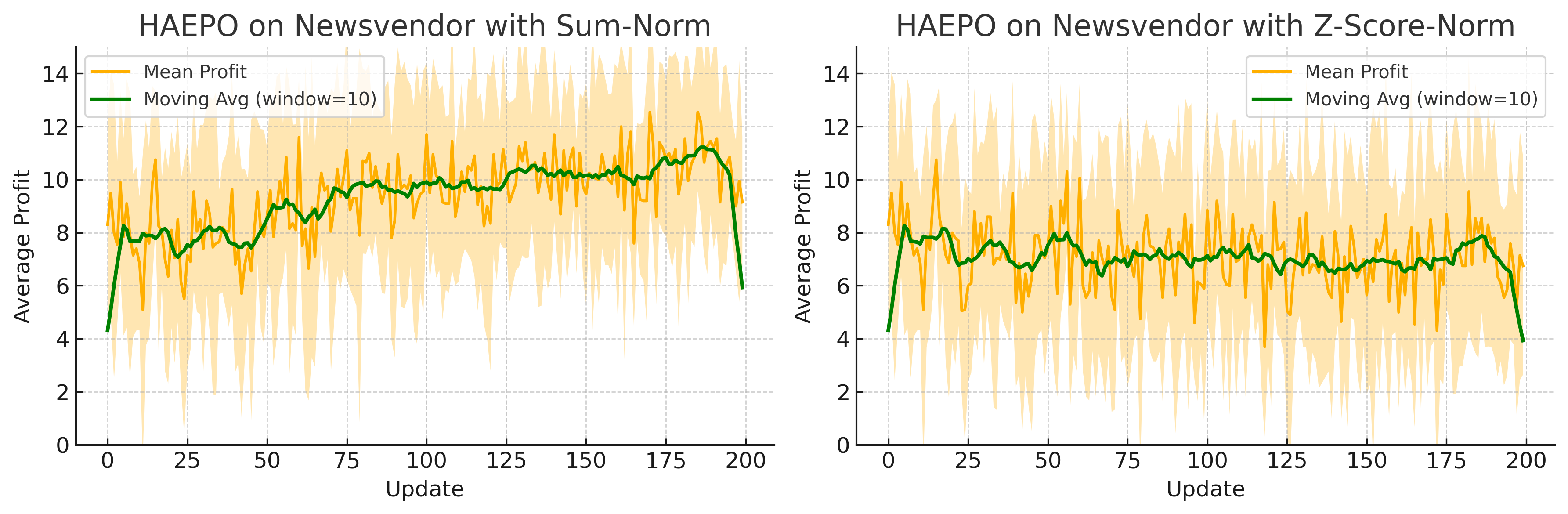}
  \caption{HAEPO on the Single-Period Newsvendor: sum-norm vs.\ z–score.}
  \label{fig:newsvendor}
\end{figure}

\subsubsection*{Experimental Setups}
\begin{description}
  \item[Chain MDP]  
    We evaluate on a deterministic 5-step “chain” environment \citep{sutton2018reinforcement}.  
    At each time-step $t=0,\dots,4$, the agent chooses 
    $a_t\in\{\text{advance},\text{stay}\}$.  
    \begin{itemize}
      \item If $a_t=\text{advance}$, state $s_t\to s_{t+1}$; otherwise $s_t$ remains unchanged.  
      \item Only upon reaching $s_5$ does the agent receive a terminal reward $R=1$, all other trajectories yield $R=0$.  
      \item \textbf{Horizon:} $T=5$ steps.  
      \item \textbf{Action space:} 2 discrete $\{\text{advance},\text{stay}\}$.  
      \item \textbf{Batch size:} $M=8$ full episodes per update.  
      \item \textbf{Learning rate:} $\alpha=0.1$.  
      \item \textbf{Updates:} $200$ gradient steps.    
    \end{itemize}

  \item[Newsvendor]  
    We use the canonical single-period inventory (“newsvendor”) problem \citep{Arrow1951OptimalIP}  
    as a one-step decision benchmark.  
    \begin{itemize}
      \item Agent selects order $q\in\{0,1,\dots,10\}$, then observes demand $d\sim\mathrm{Poisson}(5)$.  
      \item Profit per episode: 
      \[
        R(q,d) = p\min(q,d) - c\,q + v\max(q-d,0),
      \]
      with $p=10$, $c=6$, $v=2$.  
      \item \textbf{Horizon:} $T=1$ step.  
      \item \textbf{Action space:} 11 discrete order levels.  
      \item \textbf{Batch size:} $M=8$ samples per update.  
      \item \textbf{Learning rate:} $\alpha=10^{-3}$.  
      \item \textbf{Updates:} $200$ parameter updates.   
    \end{itemize}
\end{description}

\subsubsection*{Theoretical Expectations}
\begin{itemize}
  \item \emph{Sum–Norm:} Bounded rescaled returns $\tilde R_k = R_k / \sum_j R_j$, stable in dense‐reward regimes.  
  \item \emph{Z–Score:} Whitened returns $\tilde R_k = (R_k - \mu)/\sigma$, unbiased with lower variance in sparse or dispersed‐reward regimes.
\end{itemize}

\subsubsection*{Results}
\begin{itemize}
  \item \textbf{Chain MDP (Fig.~\ref{fig:chain-mdp}).}  
    Z-score normalization yields a markedly faster and more stable learning curve. The mean episodic return climbs to about 0.8 by the 100th update and remains tightly concentrated across seeds (low variance). In contrast, sum-normalization fails to reliably credit the sparse end-of-chain reward, plateauing around a mean return of 0.5 and showing large fluctuations between runs-evidence that without whitening, credit assignment in long‐horizon tasks can be both slow and noisy.
  \item \textbf{Newsvendor (Fig.~\ref{fig:newsvendor}).} The normalization of the sum preserves the magnitude of the one‐step profit signal, driving the mean profit to approximately 10 units within the first 50 updates and maintaining very narrow confidence bands between seeds. In contrast, z-score normalization compresses the modest profit variations inherent in this task, causing slower progress (peaking around 7 units) and producing noticeably wider error bars, which is evidence that the retention of the sum norm in absolute scale yields faster and more stable learning in dense settings of low variance.
\end{itemize}

These analyses confirm that tailoring the reward normalization to task horizon and reward dispersion yields the most robust and stable HAEPO updates.

\section*{Human Evaluation Form}
\label{appendix:human-eval}

This appendix section shows the exact questionnaire of a seed presented to each participant when collecting human preference ratings for TL;DR summaries generated by GRPO \citep{shao2024deepseekmath}, DPO \citep{Rafailov2023DirectPO}, and HAEPO for TL;DR \citep{volske-etal-2017-tl}. We intended to reach out to 60 participants for the preferences, but ended up with 40. We reached out to our participants using social media and secondary connections. Each seed had 8 participants. 

Each seed had a \emph{randomized} sequence of six summaries: two of each algorithm (HAEPO, DPO, GRPO), generated by two backbone models (Qwen 2.5, 1.5 B parameters; Llama 3.2, 1 B parameters). They did not know which system produced which summary.

In Figure~\ref{fig:human-eval-form}, the six model‐generated summaries are shown in a fixed sequence for each Reddit post. 
For a given Reddit Post, the combinations of algorithms and models are in the table~\ref{tab:reddit}.

\begin{table}[H]
  \centering
  \scriptsize
  \setlength{\tabcolsep}{3pt}
  \caption{Autocompletion slots (1--6) for each Reddit post.}
  \label{tab:reddit-autocomplete}
  \begin{tabular}{@{}cll@{}}
    \toprule
    Slot & Reddit Post 1 & Reddit Post 2 \\
    \midrule
    1 & DPO on \textit{Llama\,3.2}  & HAEPO on \textit{Llama\,3.2} \\
    2 & GRPO on \textit{Llama\,3.2} & GRPO on \textit{Llama\,3.2} \\
    3 & HAEPO on \textit{Llama\,3.2} & DPO on \textit{Llama\,3.2} \\
    4 & GRPO on \textit{Qwen\,2.5}  & GRPO on \textit{Qwen\,2.5} \\
    5 & HAEPO on \textit{Qwen\,2.5} & HAEPO on \textit{Qwen\,2.5} \\
    6 & DPO on \textit{Qwen\,2.5}   & DPO on \textit{Qwen\,2.5} \\
    \bottomrule
  \end{tabular}
  \label{tab:reddit}
\end{table}

\clearpage
\onecolumn
\setlength{\fboxsep}{6pt}
\scriptsize
\noindent\fbox{%
  \begin{minipage}{\textwidth}
    \subsection*{Instructions}
  
    Welcome to this small questionnaire. You will be given two Reddit posts. You have to read the post and rate the following 6 summaries on the scale given below.
    \[
      \begin{aligned}
        1 &= \text{Poor},       &\quad 2 &= \text{Fair},       \\
        3 &= \text{Good},       &\quad 4 &= \text{Very Good}, \\
        5 &= \text{Excellent}.
      \end{aligned}
    \]
    \subsection*{Reddit Post 1}
    \begin{quote}
    \textbf{SUBREDDIT:} r/relationships\\
    \textbf{TITLE:} Me [20F] and my s/o [20M] of nine months, got into our first fight. How do i deal with this?\\[0.5ex]
    \textbf{POST:} Backstory about myself before I begin: This is my first real relationship. I've never really dated because I had never really seen the point. I am now dating my best friend. We never really fight but instead just talk through our problems any time we have them. We both have anxiety issues but I still have not been able to conquer mine…

    Long story short, my boyfriend and I got into our first real fight over something I've done (about 2 months ago). I can confidently say that I royally fucked up. I lost his trust. He almost broke up with me a few days before Christmas but we decided to try to fix the problem instead.

    The best way to explain what I've done without saying too much is saying that he told me something very personal and it scared me. I went to a friend (who I thought I could trust) for advice on the problem because I was genuinely scared and didn't know what to do. (It was one of those situations where I felt that I couldn't go to him to tell him how I felt nor could I figure out how I felt at the moment.) Before I could tell him that I had told her, she got mad at me over something stupid and told him that I told her.

    Basically what I'm asking is, how do I go about gaining his trust again and proving that I love him and that I want to be with him?
    \end{quote}

    \subsection*{Responses: }
    \begin{itemize}
      \item \textit{Got into first fight with boyfriend over something that I've done. Lost trust in him and he told a friend about it. Now he's mad at me because he thinks I've betrayed his trust. How do I fix this?} 
      \item \textit{I fucked up my relationship with my boyfriend and he broke up with me. How do I regain his trust in me?} 
      \item \textit{My boyfriend and I got into our first fight because I told him something that he thought I was lying about and he almost broke up with me. How do I fix this?} 
      \item \textit{My boyfriend and I got into our first fight. How do I prove that I love him and that I want to be with him?} 
      \item \textit{Boyfriend and I got into our first fight. How do I prove that I love him and that I want to be with him?} 
      \item \textit{First real relationship. Got into first real fight. Worried about losing his trust. Worried about being unfaithful. Worried about breaking up. How do I fix this?} 
    \end{itemize}
    \subsection*{Reddit Post 2}
    \begin{quote}
    \textbf{SUBREDDIT:} r/books\\
    \textbf{TITLE:} Is it going to get better?\\[0.5ex]
    \textbf{POST:} I've been pretty slack in terms of reading novels for say the past 10 years (except for Harry Potter), but decided over the last few months to get over myself and get back into reading.

    I've got a list of good books to read (no idea where from) and decided to just buy book titles based on the title alone.

    I read 1984 (pretty awesome) and then read The Wind Up Bird Chronicle (I finished both of them!).

    I enjoyed both of them, my only problem being that I am struggling to read them and maintain interest for prolonged periods of time. I will read 2 pages and get exhausted, or read 3–4 pages and realise I haven't taken anything in.

    Is this something that is going to improve? At the moment it feels like reading is a chore as opposed to a leisure activity.

    The next book on my list is \emph{Do Androids Dream Of Electric Sheep}.
    \end{quote}

    \subsection*{Responses: }
    \begin{itemize}
      \item \textit{Decided to rekindle my love for reading after a decade-long slump. Have trouble maintaining interest in reading. What can I do to help improve this?} 
      \item \textit{I'm getting back into reading after a decade long break. What are some tips to improve reading speed and increase enjoyment of books?} 
      \item \textit{Decided to rekindle my love for reading after a decade long break. Have started with two great books and feel like a chore. What are some tips to improve reading speed and enjoyment?} 
      \item \textit{I read 2 pages of "The Wind Up Bird Chronicle" and realised I don't enjoy it. Is it something that I will get into the next decade?} 
      \item \textit{I read 2 pages and realised I don't enjoy reading. Is it something that will improve?} 
      \item \textit{I've been reading a bunch of stuff for the past few months and have realised that I don't have the patience for long novels. I'm looking for something that I can finish in a couple of hours and still enjoy. Is this something that will improve?} 
    \end{itemize}
  \end{minipage}
}
\captionof{figure}{Human preference questionnaire for one seed, showing the original Reddit posts and the six anonymized summaries (two per algorithm, two backbone models).}
\label{fig:human-eval-form}

\end{document}